# Markov Cricket: Using Forward and Inverse Reinforcement Learning to Model, Predict And Optimize Batting Performance in One-Day International Cricket

Manohar Vohra[1] and George S. D. Gordon[1]

[1] University of Nottingham, Nottingham NG7 2RD, United Kingdom
`{eexmv4,george.gordon}@nottingham.ac.uk`

**Abstract.** In this paper, we model one-day international cricket games as Markov processes, applying forward and inverse Reinforcement Learning (RL) to develop three novel tools for the game. First, we apply Monte-Carlo learning to fit a nonlinear approximation of the value function for each state of the game using a score-based reward model. We show that, when used as a proxy for remaining scoring resources, this approach outperforms the state-of-the-art Duckworth-Lewis-Stern method used in professional matches by 3 to 10 fold. Next, we use inverse reinforcement learning, specifically a variant of guided-cost learning, to infer a linear model of rewards based on expert performances, assumed here to be play sequences of winning teams. From this model we explicitly determine the optimal policy for each state and find this agrees with common intuitions about the game. Finally, we use the inferred reward models to construct a game simulator that models the posterior distribution of final scores under different policies. We envisage our prediction and simulation techniques may provide a fairer alternative for estimating final scores in interrupted games, while the inferred reward model may provide useful insights for the professional game to optimize playing strategy. Further, we anticipate our method of applying RL to this game may have broader application to other sports with discrete states of play where teams take turns, such as baseball and rounders.

## 1 Introduction

Cricket is the second-most popular sport worldwide in terms of fan-following [1]. It is a game played between two teams with each taking one turn to score points, termed 'runs'. The first team aims to score as many runs as possible before their turn (an 'innings') is over by hitting the ball (termed 'batting') while the second team try to limit scoring opportunities by how the ball is delivered to the batter (termed 'bowling'). The batting team begins an inning with 10 'wickets', where the innings can be ended when these wickets are depleted (e.g. through the ball being caught or the ball hitting the stumps) or else when the ball has been bowled a maximum number of times (counted in groups of 6 bowls called 'overs'). After the first team's innings is complete, the roles are reversed and the second team tries to score a higher number of runs in order to win. In the one-day format of the game, which this study focuses on, each team has one innings with 50 overs, representing 300 bowls of the ball. The balance between scoring aggressively while not losing wickets is what makes the game exciting.

Each bowl of the ball in cricket can change the score, the number of wickets remaining and the number of balls left. The game is therefore well-suited to being modelled as a Markov process moving semi-randomly between states of the game, with each bowl of the ball representing a transition between states. This Markov model enables the use of *reinforcement learning* to infer key parameters about the game.

Reinforcement Learning (RL) is a branch of machine learning which utilizes the psychological concept of learning via reinforcements [2]. One can train the machine or agent by providing an environment, described using states, and a set of actions to either explore or exploit (i.e. based on given optimal actions). While doing so, the agent accumulates rewards and computes future expected returns in taking the given action in that particular state. Eventually, the agent is able to behave optimally, gathering the maximum rewards possible. This methodology has been applied to various applications for both prediction and control, such as robotics [3], autonomous vehicles [4], playing games [5], crop yield prediction [6], network traffic prediction [7] and more [8–11].

The contribution of this work is three-fold. First, we devise and fit a Markov model that serves as an alternative to the DLS method using RL techniques. Second, we apply inverse RL to obtain the true reward function which can be used to provide optimal game strategies. Finally, we present a game simulator based on Markov Decision Processes (MDP) that produces a posterior probability distribution of final scores from any state of the game.

## 2  Related work

Previous methods of modelling cricket games were developed to estimate final scores or targets due to game interruptions. In general, they model the scoring resources each team has remaining and use this to adjust the target for the other team or the final scores. The game of cricket is sensitive to environmental conditions it is being played in. A game can be interrupted due to unplayable conditions such as rain or bad lighting [8]. If the game is shortened, there have been numerous methods proposed to make it fair for both teams upon resumption as one team might have received more resources to score runs. More often than not, the second innings has fewer overs to bat in. Currently, the Duckworth-Lewis-Stern (DLS) method is used to revise the final score of the team batting first to subsidize the imbalance in resources. It may also be used to estimate the final score of the team batting second in case the game is called-off and the winner is to be decided [9, 10]. The method is widely adopted in almost all top-tier cricket games and utilizes two resource parameters, overs and wickets, in a model based on completed match data. Essentially, it converts a two-dimensional resource problem into a one-dimensional resource percentage. However, a recent study has indicated that the method provides a statistically significant advantage to one team over the other [11]. This is possibly due to the model being underfitted as it uses very few parameters.

There have also been methods which look at the probability of winning the game for each team and have aimed to maintain this probability across the interruption [8, 12, 13]. Although this is a superior concept than the resource-based mechanisms, the methods have not been adopted by cricketing bodies, with DLS remaining the international professional standard. However, there have been numerous studies that have identified significant shortcomings and biases of the DLS method [1, 11, 12, 14, 15].

There have been many attempts to find alternative methods however, shortcomings in particular scenarios have made it unsuccessful in replacing the DLS method. For instance, the VJD method had inconsistencies upon resumption [15]. The method proposed by Preston and Thomas [8] utilizes only a few parameters for their probability calculations. Carter and Guthrie [12] present a method whereby more runs before interruption implies more runs to score after and vice versa. Amjad [16] had issues with long duration interruptions and unique games, which cannot easily be approximated. Similarly, Ali and Khusro [17] found numerous edge cases that were difficult to resolve. Lastly, the model presented by Hogan et al. [14] had issues with highly contested games. To the knowledge of the authors, no previous work has utilized RL techniques.

## 3 Methodology

### 3.1 Markov Reward Process

To be able to apply RL techniques, we first model a cricket game as a Markov Reward Process (MRP). For this, we must identify the reward function and states. The former is chosen to be the fraction of total runs scored in the particular state of the game. Two different state structures are defined for each innings of the game. Both model the innings in a ball-by-ball manner, increasing the granularity from the DLS method. The first innings model has a state structure with dimensions [51,11,50,6,2]. Here, the first dimension represents the over number, ranging between zero and 50. The second dimension represents the wickets lost, ranging between zero and 10. The third dimension provides information on the team's current score using bands of 10-run intervals indexed from 0 to 49. This allows us to profile various strategies based on the team's current score: for instance, if they have a low score towards the end of the game, the team might take greater risks. The fourth and fifth dimension of the state structure represents the ball within the over. The reason why we have two states for each ball is to do with the notion of extras within the game. An extra is when an illegal ball was bowled and so it has to be re-bowled. Multiple extras are handled by staying in the same state.

The state structure for the second innings has dimensions [51,11,50,50,6,2]. We have an additional dimension which takes into account the target score band, i.e. the final score of the first innings team, represented in the same manner as the current score band. This is a vital piece of additional information that allows us to profile different strategies. For example, if a team has to chase a high target, they might approach the game with higher initial scoring rate than compared to a low target. The second innings state structure is illustrated in Fig. 1.

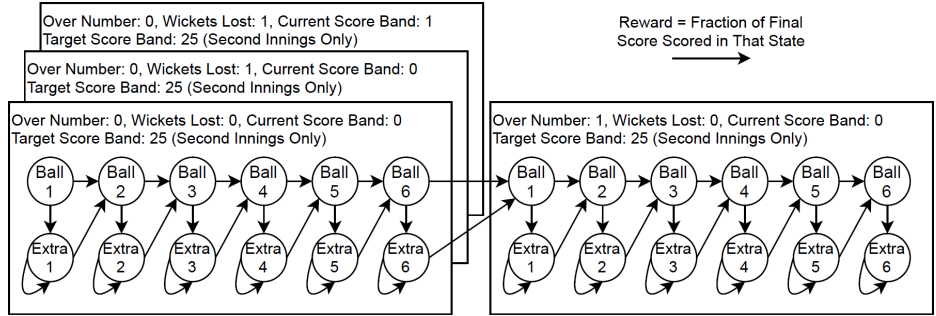

**Fig. 1.** Part of the MRP for the ball-by-ball model. The diagram indicates how the transition takes place between overs where the wickets lost or current score band does not change. The layers behind the first over block signifies how the state can change from any ball into losing a wicket or an increment in the current score band.

It is often the case when developing such state models that the data may not strictly adhere to the Markov property. For this reason we initially chose to use a Monte Carlo (MC) learning approach which is robust against such deviations [18]. In addition, the state structures are fairly large and so value function approximation techniques are applied here. To validate these function approximations within the MC method, the entire dataset is split into training and testing. The split chosen here is 90% and 10%, respectively. Furthermore, a 10-fold cross-validation is also performed.

To test the network, a random interruption point was created after the 20th over of an innings from the testing data. The state information at this point is obtained, and model inference is performed to provide the value function which represents the percentage of scoring resources left. For comparison with the gold standard, this same percentage is also obtained from the DLS table for the relevant state of the game. Then, the following formula is used to obtain the predicted final score of the team:

$$PredictedFinalScore = \lceil \frac{ScoreAtInterruption}{1-r} \rceil \quad (1)$$

In the above equation, $r$ is the value obtained from the function approximator or the DLS table. Another test is then used to quantify how much the amount of scoring resources left matches what the team eventually ended up scoring, defined as a percentage error between predicted and actual scores:

$$\%Error = \frac{PredictedFinalScore - ActualFinalScore}{ActualFinalScore} * 100 \quad (2)$$

The percentage error is then averaged for each game in the test set and represents the mean percentage error for the particular fold. The process is repeated 10 times, with the error of the folds being averaged. The standard deviation of the mean errors across the folds are also computed. These metrics are then compared between the DLS and the Monte-Carlo learning approach.

The value function approximator we use here is a neural network with several hidden layers. The input layer has five neurons for the first innings model and six neurons for the second innings model. The output layer in both cases will be a single neuron representing the value function of the given input state. An example configuration is illustrated in Fig. 2. The number and width of hidden layers was optimised to provide minimal error and standard deviation.

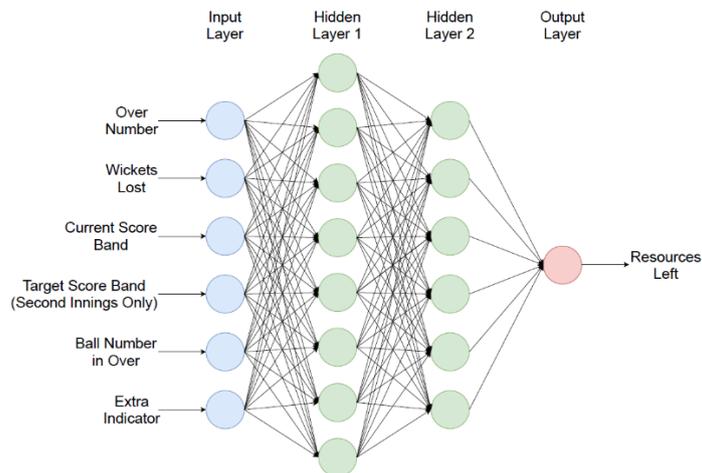

**Fig. 2.** Example neural network with two hidden layers for value function approximation of the second innings model.

In terms of usability of the proposed method, the neural network will simply replace the DLS table. A simple computer program could be developed which would take as input the state information of the game, and return as output the resources-left percentage. This could be used a direct replacement for the DLS method. Moreover, unlike the DLS implementation, this method can also be extended to provide other insights into the game. For instance, if rewards are only given for winning, the value function will represent the chance of winning instead.

### 3.2    Estimating the True Reward Function and Obtaining the Optimal Actions

In the MRPs described in the previous section, the reward function was based entirely on points scored. However, this does not reflect the true complexities of the game – for example scoring points in different ways (e.g. hitting the ball high in the air, aiming to get a 'six') carries different levels of risks of losing a wicket which reduces their average reward. Further, MRPs model the game as a series of stochastic transitions between states, which also does not reflect reality as a team can make decisions about how to play – they can choose an *action* to take in each state, though there is only a certain probability that this action will be successfully executed. Adding actions to this game model creates a Markov Decision Process (MDP). Using Inverse Reinforcement Learning (IRL), an estimate of the true reward function for this MDP of a cricket game can be formed by examining the optimal set of actions, a 'strategy', used by teams. Here, IRL techniques have been applied for the first innings only. The IRL method employed was proposed by Ng and Russell [19], but may be considered a form of guided-cost learning [20]. The main distinction comes from how the optimization problem is resolved. Further details on the approach are provided in Algorithm 1.

---
**Algorithm 1:** Inverse Reinforcement Learning (IRL) algorithm used
---
**Result:** Coefficients or weights for each feature
Define a set of features;
Compute the expert value function for each feature;
Initialize all coefficients to 1;
**while** *There are non-expert trajectories remaining* **do**
    Compute the non-expert value function for each feature using the set coefficients;
    Solve the optimization problem using linear programming based on pool of conditions plus a new condition;
    Update coefficients based on optimization results Add the conditions to the pool of conditions
**end**
---

The approach first requires a set of features to be defined. These are then combined in a linear reward function. For the first innings MRP defined in the previous section, these features are set to be the five variables used to describe the state. On top of this, to be able to determine the optimal actions to take, the reward function must not only be a function of the state, but also of the actions that can be taken by the batters. The actions include, to not score any runs (a dot ball), or to score one, two, three, four, or six runs. Each of these will be input as a feature to the reward function with an exception for dot balls. This is mainly due to the fact that it would not provide any additional reward over the basic reward function based on the state-features only as dot balls do not result in score increases or wicket losses. The value of these actions are multiplied by 1000 to further amplify their differences. Moreover, each feature will have a coefficient which is the parameter that is optimized in the algorithm. The final reward function takes the form:

$$r(s,a) = \begin{cases} x_1 s_1 + x_2 s_2 + x_3 s_3 + x_4 s_4 + x_5 s_5, \text{if } a = 0 \\ x_1 s_1 + x_2 s_2 + x_3 s_3 + x_4 s_4 + x_5 s_5 + y_a * 1000 * a, otherwise \end{cases} \quad (3)$$

Next, we must define expert and non-expert strategies. Here, we select expert strategies to be those set of actions adopted by the winning team in any game, and non-expert strategies to be those adopted by losing teams. There is often an element of chance in winning a game and so it can be lost despite using an optimal strategy, and it is also possible to win using a sub-optimal strategy, provided it is more optimal than the opposing team. However, when averaged over many games we expect that maximizing the difference in rewards between winning and losing teams will produce a suitable approximation of the true reward function.

Finally, we bring this together to solve the following optimization problem, based on that proposed by Ng and Russell [19]:

$$maximize \sum_{i=1}^{k} p(\hat{V}^{\pi^*}(s_0) - \hat{V}^{\pi_i}(s_0)) \quad (4)$$

where $V^{\pi*}$ is the optimal strategy for a sequence of states $s_0$, $V^{\pi}_i$ is suboptimal strategy *i,* and *p(..)* represents a distance metric, which is here defined to be Euclidean distanced squared. In other words, we are maximizing the distance between the optimal strategy and all suboptimal strategies.

### 3.3 Game Simulator

For the final piece of software, the aim was to create a complete MDP modelling a cricket game and using this to simulate games. The reason this will be a complete MDP is because of the inclusion of transition probabilities, which takes into account the random behavior when an action is chosen but is not executed. For instance, the batter may decided to hit the ball over the boundary to obtain 6 runs, but the ball may instead be caught and a wicket lost. When applying MC learning as in the previous section, the actions are pre-determined for a given game and so the observed transitions are assumed to include transition probabilities. In addition, the data was also assumed to be generated using an optimal policy. However, the improved estimate of the reward function obtained using IRL allows us to compute the optimal policy and thus infer the transition probabilities from the dataset. With an optimal policy, reward function and transition probabilities, the MDP is now complete.

At first, the conditional probability of scoring a particular amount of runs (i.e. between zero and six) for all visited states, given that it was not a wicket ball, was computed using historic data. Along with this, the probability of a wicket falling, given the state of the game, was also calculated. This was done separately for the first and second innings, with the ball-by-ball state structure being used. For the first innings, the probability of selecting the non-optimal action was also acquired.

From there, a game can be started from the initial state. Based on the probability, it is first identified whether or not a wicket falls in this state of the game. If not, the probability of not selecting the optimal action is used to determine whether the action selected should be optimal or not (done only for the first innings). If not, the action, which represents the number of runs scored, is selected randomly based on the respective transition probabilities in the current state. After all this is known, the next state of the game can be computed. The simulator then goes on to check whether the innings has ended, otherwise it transitions to the next state. Once the first innings of the game is completed, the second innings is simulated in a similar manner.

The simulator can then be used to generate samples from a posterior probability distribution of possible final scores when simulated 100 times. Games were interrupted in the same manner used for testing the proposed method in Section 3.1, but in this case the entire dataset was used. The error percentage was computed once again to compare the average simulated score to the actual score of the team. These error percentages were then averaged and the standard deviation computed to quantify the performance of the simulator. This methodology was also repeated for the second innings of the game.

## 4 Results and Discussion

### 4.1 Alternative to the Duckworth-Lewis-Stern Method

The DLS table has been shown for reference in Fig. 3a. Fig. 4 provides the various neural network configurations that were examined for the first innings model. The net configuration was varied with the number of neurons in the hidden layers and the number of epochs used for training.

In general, the strategy used was to increase the number of neurons to improve the percentage error, while increasing the epochs to increase the stability of the model with the reduction of the standard deviation. Deeper nets were also tested to improve performance as it would be able to model additional complexities in the data. To avoid a lucky train-test split, 10-fold cross-validation was used. This is the reason there are slight variances in the metrics for the DLS which are not changed in any way other than through selection of testing data.

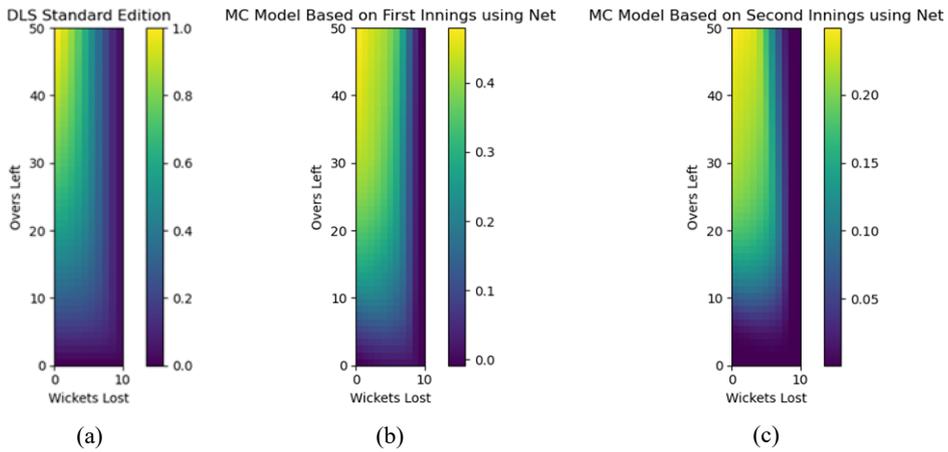

**Fig. 3.** (a) The DLS standard edition table. This holds the percentage of scoring resources left, shown as a ratio, given the number of wickets lost and the number of overs left. (b) The value function of the proposed first innings model averaged into an over-by-over model for comparison only. (c) The value function of the proposed second innings model averaged into an over-by-over model for comparison only.

After computing the mean and standard deviation of the percentage error across all folds, the net with the best performance was configuration number 6. The range of error is between -2.11% and -0.11%, which has clearly outperformed the DLS method, which has an average error between 2.77% and 5.41%. As seen by these statistics, the proposed model underestimates the scoring resources, while the DLS method overestimates with a larger magnitude.

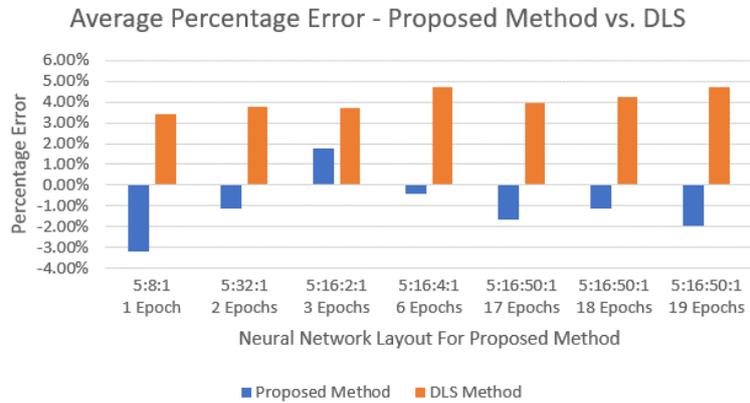

**Fig. 4.** The various net configurations and their cross-validated error percentages as tested for the first innings model. Cross-validation ensures repeatability and reliability of the experiments.

In addition, to compare the proposed model with the DLS method visually, the value function of the states have also been averaged into an over-by-over model (i.e. averaging the 50x6x2 layers of the 51x11x50x6x2), shown in Fig. 3b. Although the values are almost half of those in the DLS table, this may be due to there being many unvisited states in the model and so the error in prediction, especially given that it was averaged, would be accumulated. However, the trend from this does have similar properties to the DLS table.

The above steps were also repeated for the second innings model with its own state structure. The neural networks which were tested, with their performances, are shown in Fig. 5. The net configurations were altered with the same strategy as before. Nets with a single hidden layer struggled to reduce the error below one percent. Hence, deeper nets were tested as well.

The best performance was seen in the 8$^{th}$ test, with an average error range between -0.96% and 1.82%. This, compared to the average error range of the DLS method, which is between 2.06% and 4.54%, proves that the proposed method is able to model the amount of scoring resources left more accurately. The averaged value function is also shown in Fig. 3c. Again, the raw values of states is reduced due to the second innings model rarely using the target score bands of less than 20 (i.e. target of 200 runs). Once again, the trend does have resemblance to the first innings model and the DLS table. One noticeable difference is towards the end of the games, the resources left have a steeper reduction, indicating that the losing wickets towards the end has major drawbacks.

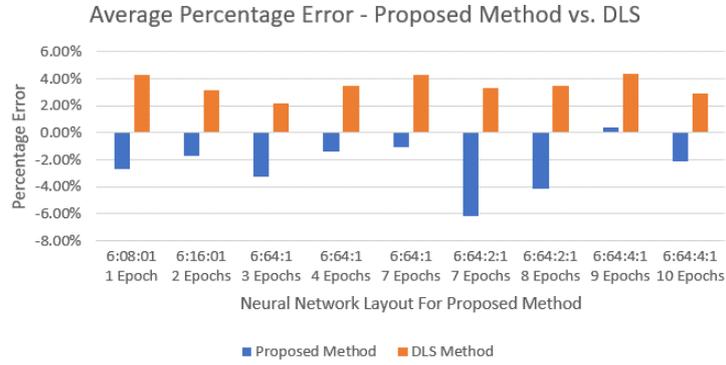

**Fig. 5.** The various net configurations and their cross-validated performances as tested for the second innings model. Cross-validation ensures repeatability and reliability of the experiments.

### 4.2 True Reward Function and Optimal Game Strategy of a Cricket Game

Fig. 6 illustrates the feature coefficient values produced using our IRL model. State features one, three, four, and five have a value of one, which represents the upper bound of the coefficients in the optimization, and so these values may been much larger if the bounds were increased. The reason why these values are high is due to the fact that they correlate to the length of the game. For instance, if many overs are played there are more opportunities for scoring. The third state feature represents the current score band of the team, and indeed, if this value is large, the team have scored many runs. Hence, it is logical for its coefficient to be large as well. To avoid excessively large weights of these features, we chose to keep the upper bound set to 1. The only feature with a negative coefficient is the number of wickets lost. As expected, losing wickets is an extremely bad outcome, and so the rewards for this should be negative.

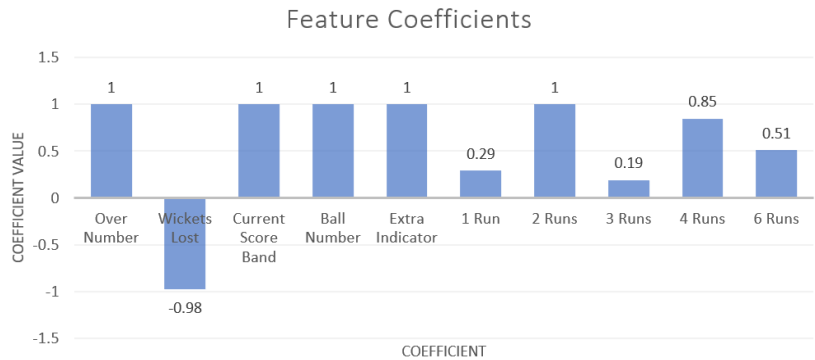

**Fig. 6.** The value of the coefficients for their respective feature.

Within the action-feature coefficients, a range of different values are observed. Scoring two runs has a coefficient value of one. The reason why this is high is perhaps because it allows the same relatively successful batter to attempt to score runs again immediately, rather than swapping with his batting partner if an odd number of runs is scored. Further, the risk of being run out is relatively low given the time typically taken for the ball to reach the wickets – this represents a good risk-reward trade-off between point scoring and losing wickets. In contrast, the coefficient for scoring a three has the lowest value among the actions, possibly because the time taken to do this comes with higher risk of being run out. Scoring a single run has a relatively low coefficient, possibly due to the relatively low rewards it provides balanced against the risk of running out of scoring opportunities.

The second-highest coefficient value among the actions is of scoring a four, which may be due to the balance it provides between accumulating many runs with low risk of getting out. However, such long hits of the ball have higher risk of being caught and so this may be why the coefficient is lower than for 2 runs. Hitting a six always involves a large risk for the batting team as the ball must be high in the air, but it does provide a large reward. Hence, the coefficient for it is just above the 0.5 mark.

The use of a linear reward model means there are only simple offsets between the full set of rewards for different states – for example, scoring 4 runs can incur different rewards in different states, but will always have the same relative value compared to scoring a 6 in any given state. A more advanced analysis conducted in future could examine a nonlinear approximation that allows the difference between rewards within a state to vary – for example, it may be more advantageous to try to score 6 on the last ball of game, whereas scoring 4 is the better option in most states.

In order to obtain the optimal game strategy, the state-action value function was computed. The trends observed in the value functions indicate that games are won when teams lose fewer wickets and play their entire quota of 50 overs. Once again, this trend remains the same for all actions due to the linearity of the reward function. Additionally, it was also deduced that dot balls play an important part during the beginning of the game and when many wickets have fallen. This is so that the team can play a longer game eventually accumulating more runs or, the fact that the batting team are being outplayed by the bowling team completely. Similarly, taking singles at the beginning also has a very high value, possibly because the strongest batters are often used first and so the bowling team uses their strongest bowlers, making scoring a 1 good trade-off between the need to score and reducing the risk of losing a top-player too early.

This theory follows with scoring a two but with smaller value. Scoring three runs, as seen before, has very few returns both, immediately and in the future. As for scoring a four, it provides the highest value among the actions. This does decrease when many wickets fall due to the risk of being all-out. Lastly, scoring a six provides higher value towards the end of the game, typically with more than half the wickets still remaining. This does have relatively smaller future returns compared to a four which is due to the high risk it involves.

### 4.3 Game Simulator

Below are some game trajectories produced by the developed simulator:

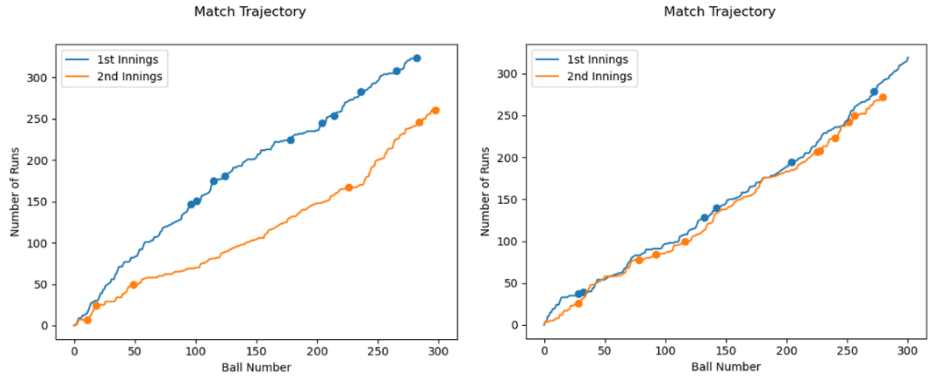

**Fig. 7.** Examples of simulated game trajectories. Dots represent points where wickets were lost.

Initially, the first innings was simulated using the inferred reward function, the associated optimal actions, and transition probabilities computed from historic data. The second innings was then simulated using only transition probabilities. In 100 simulations, the team batting in the first innings wins 82% of the time, showing the superiority of this optimal strategy. This reduces to 50% if the first innings is simulated using transition probabilities only. Nonetheless, the simulator provides realistic results: the optimal actions for most states is hitting a four, with slightly different actions recommended towards the end of the game. These actions are sometimes a one, two, or a six.

The simulator can also be used to provide a posterior probability distribution of final scores for both innings from any point in the game. Fig. 8 illustrates the distribution plots after the innings were interrupted and 100 simulations were executed from that given point. The first innings optimal action model obtained an average percentage error of 9.87%, with a standard deviation of 16.09%. The error here is very large, once again due to the optimal actions being used at times. This is expected as it is the error of the posterior distribution for a single game, compared to the previous error value which was the maximum likelihood across all games. If these are not included, the error percentage falls down to 2.83%, with a standard deviation of 15.10%. The standard deviation, on the other hand, has not changed drastically, implying that a wide range of feasible outcomes are shown by the simulator. Indeed, if the number of simulations are increased, the accuracy would also increase, but this also increases the execution time of the program.

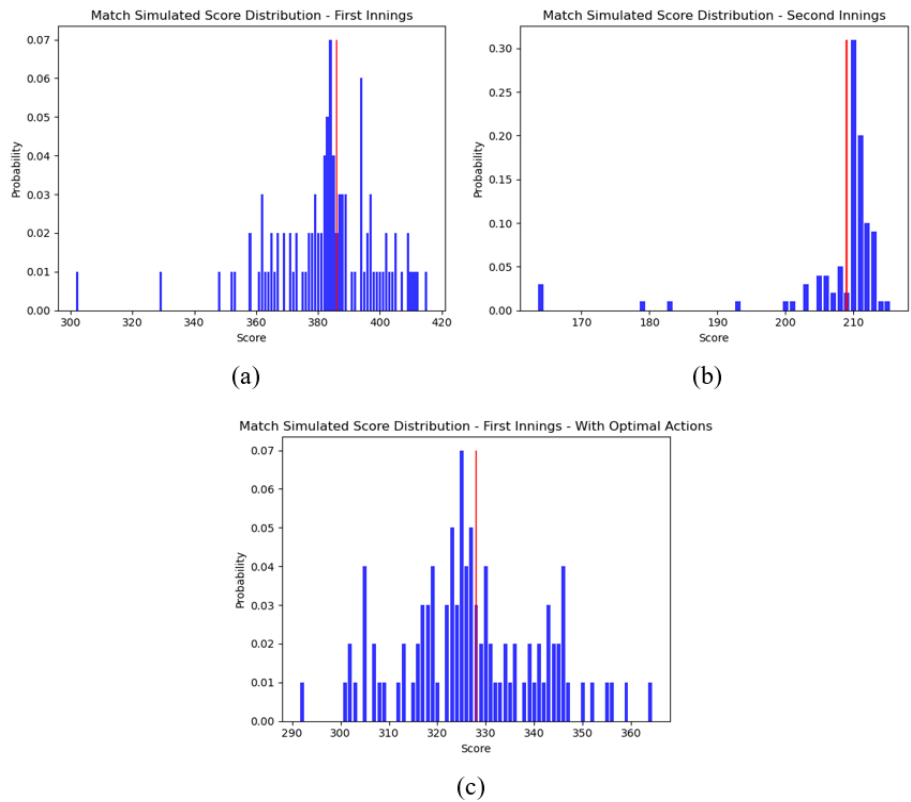

**Fig. 8.** Probability distribution of scores based on 100 simulations. The red tower refers to the actual score the team reached, while the blue are for the respective simulated scores. (a) First innings model without using optimal actions in the simulation. (b) Second innings model. (c) First innings optimal actions model, results indicate the team played optimally as with optimal actions it would have scored nearly the same.

The second innings model gave an average percentage error of 2.09%, with a standard deviation of 11.10%. This, of course, is without the optimal actions which were not acquired for the second innings. This accuracy is better than both of the first innings models, offering better stability as well. The reason for this could be due to the extra state variable which is used, providing very specific probabilities, profiling actions taken by with greater accuracy.

## 5  Conclusion and Future Work

We have presented an alternative method to the DLS method for modelling interrupted one-day cricket games that uses reinforcement learning techniques to achieve up to 10 times (First Innings: -1.11% vs 4.27%. Second innings: 0.43% vs. 4.35%) improved accuracy in determining the amount of scoring resources left. The usability of the method is similar to that of the DLS table, where the fitted neural network replaces the table. Maintaining the model is also very simple, where new match data can be used to retrain the net on-the-go. On top of this, we can utilize other reward attributes such as winning or losing, to give a more complex function. This makes the method versatile.

In addition, IRL techniques have been successfully applied to infer the true reward function for the first innings. Using this, the optimal strategy was determined, providing details of which action should be taken to maximize future returns. This information was then incorporated into an MDP, enabling full simulation of a cricket game. Simulations of this have shown realistic game projections, and an optimal strategy applied by the team batting first leads to victory on most occasions.

Finally, a game simulator was developed which provides a posterior probability distribution of the final score based on the given state of the game. Thousands of simulations can be executed, showcasing the various possible states the game can go through. On average, this is able to predict the final score with error as low as 3%.

To the authors' knowledge, this is the first time reinforcement learning has been applied to model the game of cricket, and there is significant scope to develop this approach for more detailed insights into the game. First, the current inferred reward function is a linear combination of the features, which constrains the relative value of different actions in a given state. A nonlinear approximator, such as a neural network, could be trained to allow greater flexibility. Additionally, there is other data that might be used to improve predictions for instance, the dew factor, type of pitch. These could then be included in the state variables, which may improve accuracy and improve adherence to the Markov property. Lastly, optimal actions for the second innings, which would require a different IRL approach is worth investigating given the promising results shown in this work.

**Data statement:** The ball-by-ball data for one-day international cricket matches was obtained from Cricsheet [21].